\documentclass[conference]{IEEEtran}
\IEEEoverridecommandlockouts

\usepackage{cite}
  \usepackage{graphicx}
  \graphicspath{{./figures/}}
  \DeclareGraphicsExtensions{.eps,.pdf,.jpeg,.png}
\usepackage[cmex10]{amsmath}
\usepackage{amsthm}
\usepackage{amssymb}
\theoremstyle{definition}
\newtheorem{definition}{Definition}

\theoremstyle{remark}

\usepackage{algorithm,algorithmic}

\newcommand{\eqdef}{\overset{\mathrm{def}}{=\joinrel=}}

\ifCLASSOPTIONcompsoc
  \usepackage[caption=false,font=normalsize,labelfont=sf,textfont=sf]{subfig}
\else
  \usepackage[caption=false,font=footnotesize]{subfig}
\fi
\usepackage{bm}

\usepackage{slashbox}
\usepackage{makecell}
\usepackage{arydshln}

\newcolumntype{P}[1]{>{\centering\arraybackslash}m{#1}}

\begin{document}
\bstctlcite{IEEEexample:BSTcontrol}
\title{Matching Media Contents with User Profiles\\by means of the Dempster-Shafer Theory}

\author{
\IEEEauthorblockN{Luigi Troiano}
\IEEEauthorblockA{University of Sannio\\ Department of Engineering\\ Benevento, Italy\\
Email: troiano@unisannio.it}
\and
\IEEEauthorblockN{Irene D{\'i}az}
\IEEEauthorblockA{Oviedo University\\ Computer Science Department\\ Gij\'on, Spain\\
Email: sirene@uniovi.es}
\and
\IEEEauthorblockN{Ciro Gaglione}\thanks{
\noindent
\textbf{DISCLAIMER.  
This article was prepared or accomplished by Ciro Gaglione in his personal capacity. The opinions expressed in this paper are the authors' own and do not reflect the view of Sky Italia.}
}
\IEEEauthorblockA{Sky Italia\\ Interactive Tv Lab\\ Milan, Italy\\
Email: ciro.gaglione@skytv.it}
}


%Telephone: (800) 555--1212, Fax: (888) 555--1212}


\maketitle

\begin{abstract}
The media industry is increasingly personalizing the offering of contents in attempt to better target the audience. This requires to analyze the relationships that goes established between users and content they enjoy, looking at one side to the content characteristics and on the other to the user profile, in order to find the best match between the two. In this paper we suggest to build that relationship using the Dempster-Shafer's Theory of Evidence, proposing a reference model and illustrating its properties by means of a toy example. Finally we suggest possible applications of the model for tasks that are common in the modern media industry.
\end{abstract}


% \fi
\IEEEpeerreviewmaketitle

\section{Introduction}

Digital technologies are radically changing the way of performing business in media industry, with new possibilities of tailoring the catalog so that everybody has the chance of enjoying contents that best fit his/her interests, often on demand, at the time that is most appropriate for each user. Such a change is requiring to reformulate the way of building the content  offering. Data collected from customers regarding their profile and preferences become central, so models able to interpret and to reason about data.

These models aims to discover and exploit the relationship that stands between users and media contents they enjoy. Here the problem is not to ask directly the user what are his/her interests and preferences, but to infer them by looking at those contents they access and to the feedback they provide about them. The ultimate goal is to learn a model from data able to link user to the vast catalog of contents made available by a large media company.

Looking at past interactions is useful to help users to discover contents that they would appreciate as valuable part of the product they paid for. This means to improve the customer retention and foster their upgrade towards more profitable products. The benefits coming from the implementation and use of these models go beyond existing contents and customers. They also help to propose new contents to existing customers, and on the other way to support new customers in discovering existing contents. Soon, new contents and new customers become part of the model, enriching the dataset of new entities, along a self-growing process. Predictiveness of models make them also suitable to support the acquisition of new contents and customers. 

These models are at the core logic of recommender systems (RS), that obtained large attention once Netflix showed potentiality of algorithms in developing and supporting their streaming platform \cite{Gomez2015}. Recommender systems gained large application because of the e-commerce diffusion. They are generally grouped in different types, including Content-based recommenders \cite{Salter2006}, Collaborative recommenders \cite{Candillier_comparingstate-of-the-art}, Demographic recommenders \cite{Pazzani1999}, and Hybrid recommenders \cite{Burke00knowledge-basedrecommender}.

The purpose of a recommender system is to provide a suggestion, regarding available alternatives, by scoring and ranking them according to the user preferences. In order to accomplish its task, a recommender system requires information regarding the user profile and habits with respect to the different alternatives that can be proposed to him. This information can be acquired explicitly by asking the users to rate items or implicitly by monitoring users' behavior (booked hotels or heard songs). RS can also use other kinds of information as demographic features (e.g, age, gender) or social information. The research related to RS has been focused on movies, music and books \cite{Bobadilla2013109}, being music recommendations the most studied topic, although later it has been applied to other e-commerce domains \cite{CastroSchez20112441}. 

Similar to RS, we need data about user likings regarding catalog items such as movies, series and shows. Such information can be gathered by asking the user to rate the items, e.g., by using stars or likes, or implicitly by monitoring the customer behavior, e.g., which item enjoyed fully an which partially, how often they accessed the content description, etc. In addition we need other information regarding demographics such age, gender, family members, job, etc. The objective is to relate user profiles to content descriptors. Different techniques have been experimented in order to discover and exploit this relationship. Most of them take the form of information fusion.

Following the idea explored by \cite{Zhang10}, and more concretely the model developed in \cite{Troiano201598}, we aim to build a relationship model based on the Dempster-Shafer's Theory of Evidence (D-S theory) \cite{dempster67a,shafer1976mathematical} and to use it to make inference regarding the relationship between users and contents. The reminder of this paper is organized as follows: Section II provides some preliminaries regarding D-S Theory; Section III describes the model; Section IV outlines some examples of application; Section V draws conclusions and future directions.

%In this paper, we face the problem of filtering from an e-shop catalogue a set of products which might be interesting for the customer on the basis of preferences expressed by a group of users within a market segment. This problem is studied from a theoretical point of view by means of Dempster-Shafer Theory of Evidence (D-S Theory). The purpose of this work is to show how the D-S Theory can be used in the context of RS. The first approach of this type was introduced by Zhang and Li \cite{Zhang10}.  They suggested to use D-S Theory as a mean to combine recommendations from different systems, each one considered as an independent source of information. Instead, we focus on how to use D-S Theory internally in a recommendation system. This approach goes generally with skeptism of dealing with combinations of items, which might not fit real world inventories. In this paper we propose to move from items to features in order to (i) reduce problem dimensionality and (ii) to infer user preferences even when they are not made explicit. Preferences induced by each feature are considered as an independent source of information, then combined by a rule. In the second part of this paper, it is studied how to explore the subset inclusion lattice, once Belief and Plausibility are mapped over it. We outline some efficient algorithms to perform such an exploration.

\section{Preliminaries}

The Dempster-Shafer theory, also known as the Theory of Evidence \cite{dempster67a,shafer1976mathematical}, is used as basis for the preference model presented in \cite{Troiano201598}. In D-S theory, basic probabilities are allocated to subsets, instead of elements, according to the following definitions.

\begin{definition}
A function $m:2^{\Omega}\longrightarrow [0, 1]$ over a set $\Omega$ is called a \emph{basic probability assignment} if
$$ m(\emptyset)=0 \;  and \; \sum_{A \in 2^\Omega}m(A)=1$$
\end{definition}

\begin{definition}
Let $\Omega$ be a set, then $A \subseteq \Omega$ is a focal element if $m(A)>0$. In addition, $F(\Omega) \subset 2^{\Omega}$ represents the set of focal elements induced by $m$.
\end{definition}

\begin{definition}
Let $m$ be a basic probability assignment function over a set $\Omega$. The Belief of $A \subseteq \Omega$ induced by $m$ is defined as follows
\begin{equation}
Bel(A)=\sum_{B\subseteq A} m(B)
\end{equation}
\end{definition}

\begin{definition}
Let $m$ be a basic probability assignment function over a set $\Omega$. The Plausibility of $A \subseteq \Omega$ induced by $m$ is defined as follows
\begin{equation}
Pl(A)=\sum_{B \cap A \neq \emptyset} m(B)
\end{equation}
\end{definition}

The relationship between Plausibility and Belief is given by the following equation:
\begin{equation}\label{eq:duality}
Pl(A) = 1 - Bel(\overline A)
\end{equation}
where $\overline A$ is the complement of $A$ to $\Omega$.

When the probability basic assignments are given by different sources, it is possible to combine them. The first and most common combination method is known as the Dempster's rule, that is defined as follows:
\begin{definition}
Let $m_1$ and $m_2$ be two basic probability assignments, the \textit{joint basic probability assignment} is computed as
\begin{equation}\label{eq:dempster_rule}
m_{1,2}(A) = \frac{1}{1-Z} \sum\limits_{B \cap C = A} m_1(B) \cdot m_2(C)  
\end{equation}
where
\begin{equation}
Z = \sum\limits_{B \cap C = \emptyset} m_1(B) \cdot m_2(C)
\end{equation}
is a measure of \textit{conflict} between the two basic probability assignment sets. In addition, it is assumed $m_{1,2}(\emptyset) = 0$.
\end{definition}

\begin{figure}[t!]
\centering
\includegraphics[width=6cm]{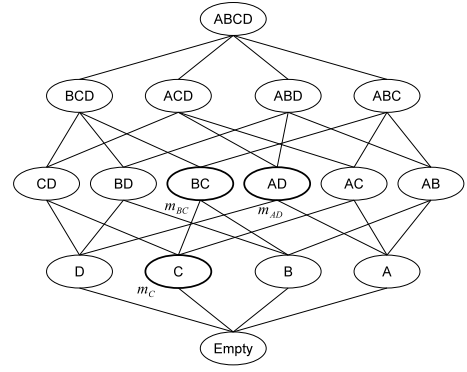}
\caption{The Boolean lattice of item subsets from $\Omega= \{A, B, C, D \}$ with focal elements $F(\Omega)=\{C,BC,AD\}$ \protect\cite{Troiano201598}}.
\label{fig:lattice}
\end{figure}

\begin{figure}[h!]
\centering
\includegraphics[height=4cm]{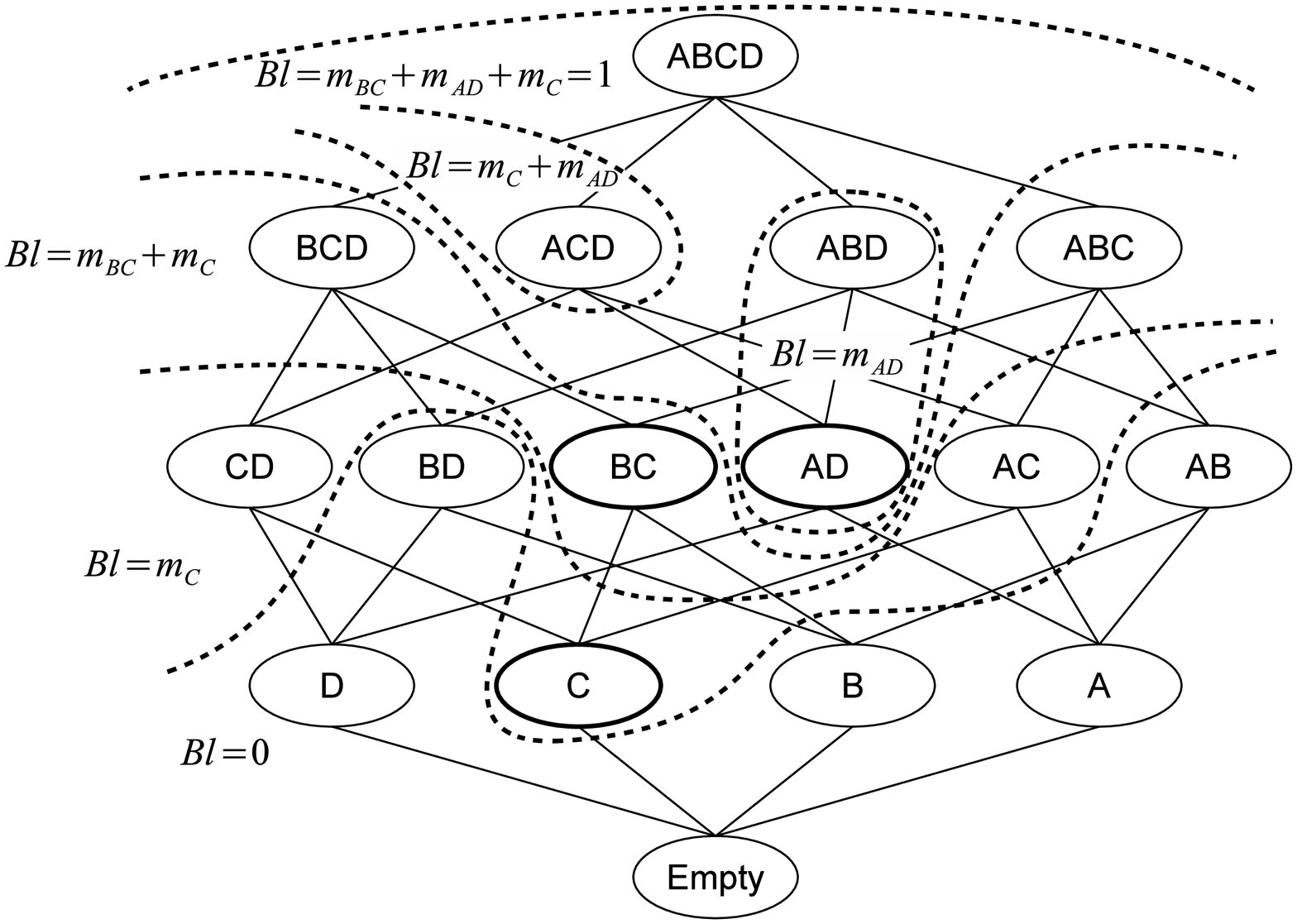}\\
\vspace{5mm}
\includegraphics[height=4cm]{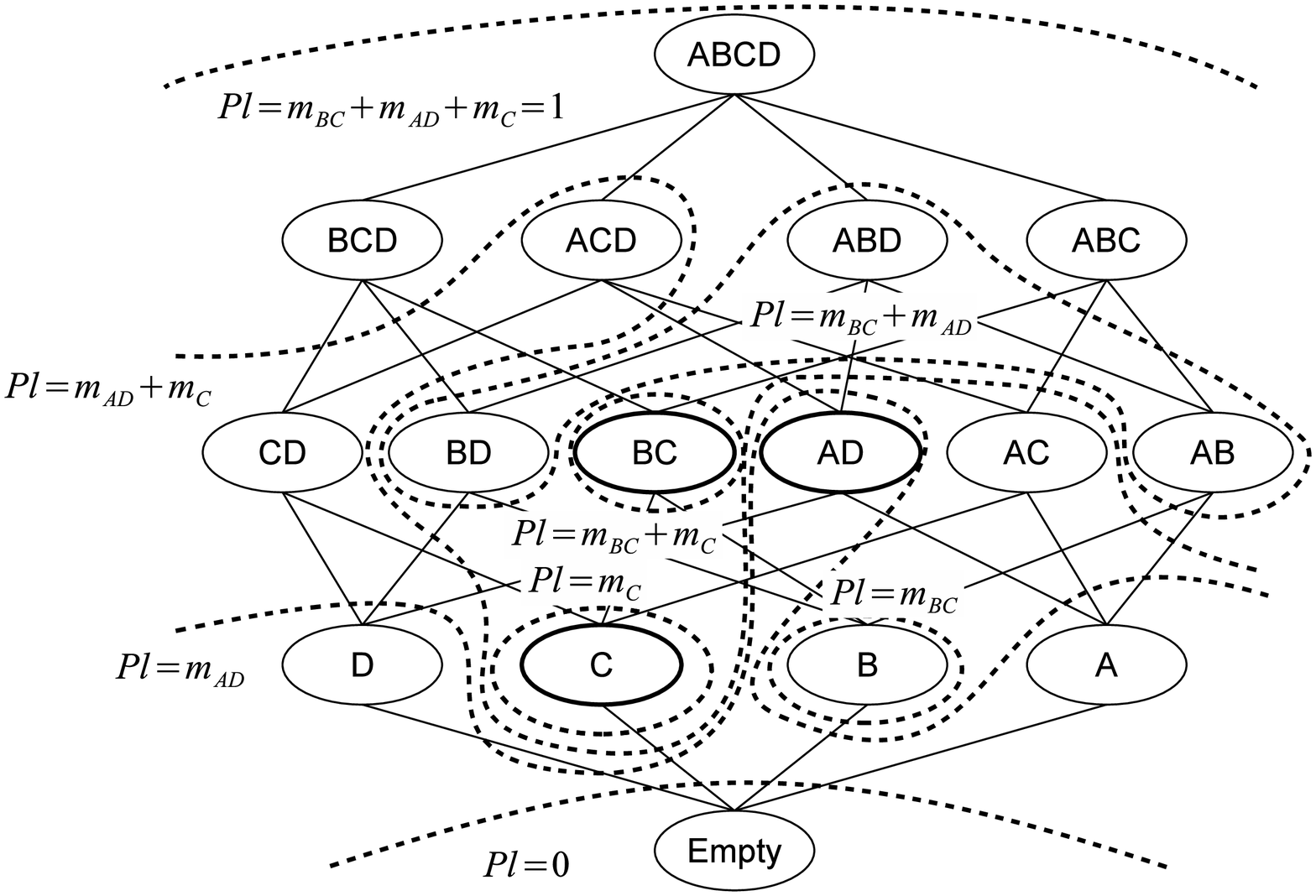}
\caption{Belief (top) and Plausibility (bottom) regions induced by $F(\Omega)=\{C,BC,AD\}$ \protect\cite{Troiano201598}}
\label{fig:lattice-bel}
\end{figure}

Belief and Plausibility are monotonic functions with respect to inclusion. This means that if we consider the lattice of $\Omega$ subsets, as shown in Fig.~\ref{fig:lattice}, Belief and Plausibility will increase from bottom ($Bel(\emptyset) = Pl(\emptyset) = 0$) to top ($Bel(\Omega) = Pl(\Omega) = 1$). In particular Belief and Plausibility will be kept constant as far as we move to nodes that do not a probability mass assigned to them. As consequence of this property, we can identify regions of connected nodes, each assuming a specific value of Belief or Plausibility, as illustrated by 
Fig.~\ref{fig:lattice-bel}.

In this example, focal elements are $C$, $BC$ and $AD$ with the associated basic probability assignments $m_C$, $m_{BC}$ and $m_{AD}$ (assuming $m_C + m_{BC} + m_{AD} = 1$). This leads to identify 8 groups in the lattice, each with Belief and Plausibility depending from a focal subset of $F(\Omega)$. Fig.~\ref{fig:lattice-bel} outlines these regions for both Belief and Plausibility. we can observe how all portions of lattice associated to a given value of Belief or Plausibility are \emph{connected}.

%\begin{figure}[h!]
%\centering
%\includegraphics[height=5cm]{lattice-plausibility.eps}
%\caption{The Plausibility areas.}
%\label{fig:lattice-pl}
%\end{figure}

\begin{figure}[t!]
\centering
\subfloat[]{\includegraphics[height=3cm]{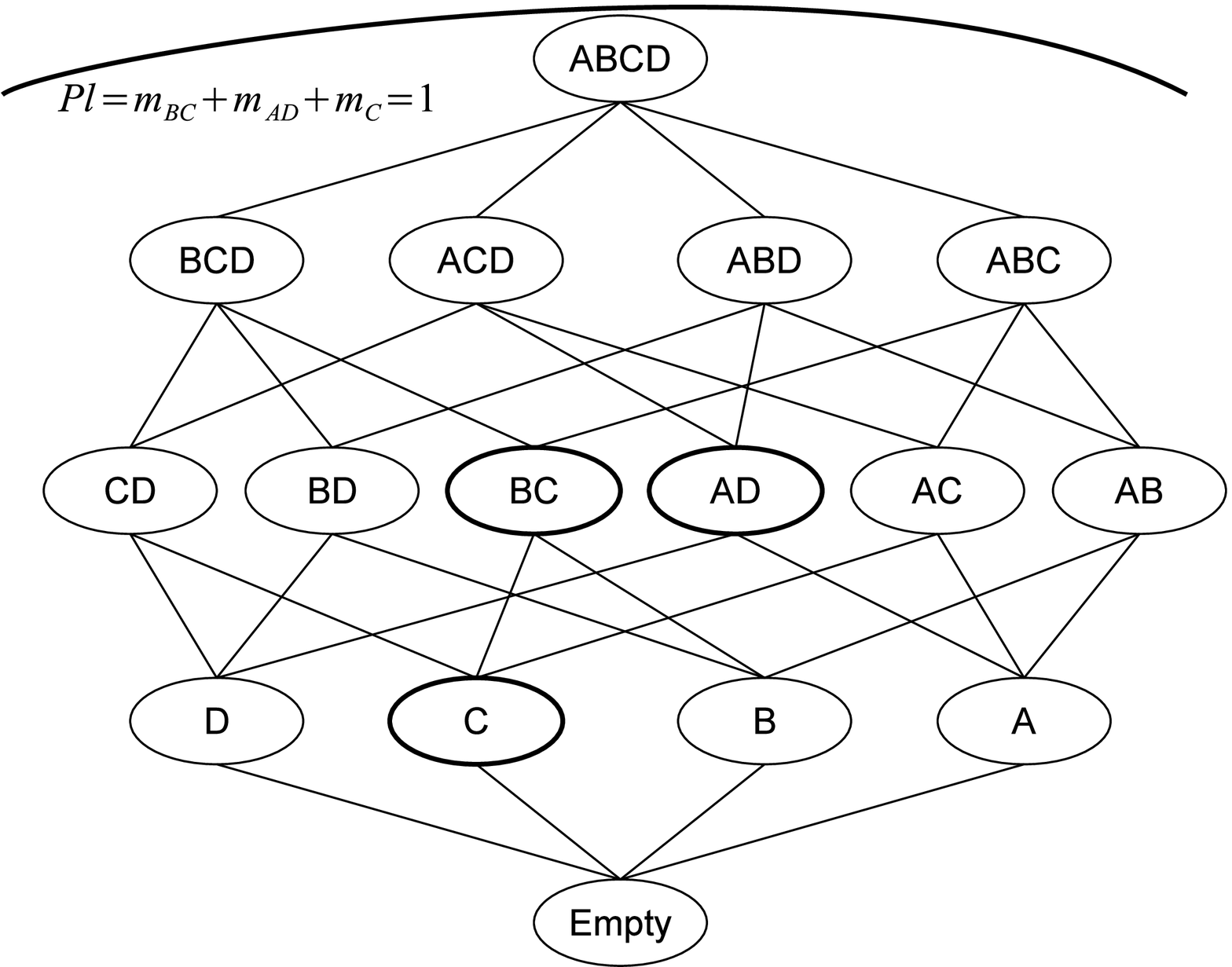}}
\hspace{4mm}
\subfloat[]{\includegraphics[height=3cm]{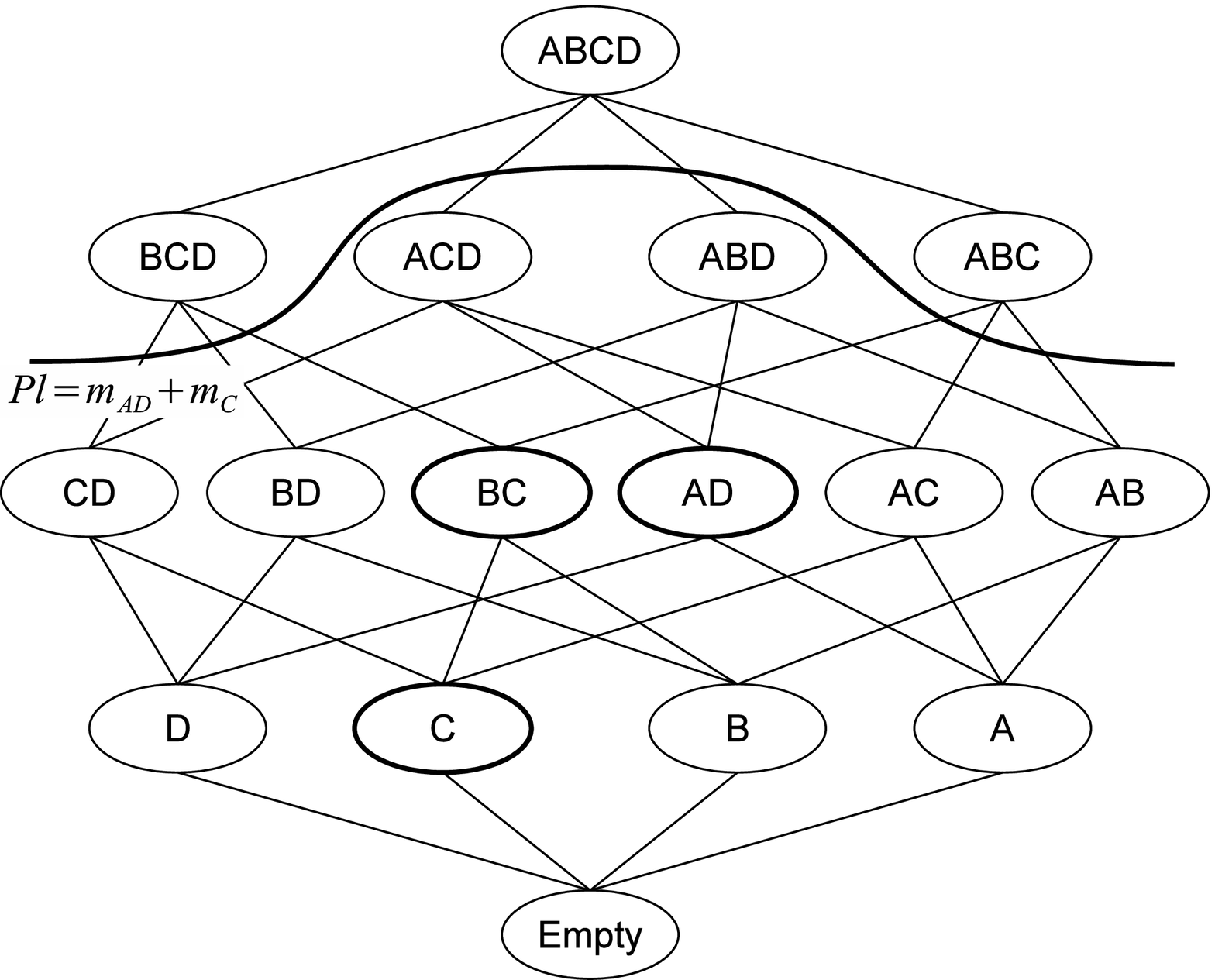}}\\
%\hspace{2mm}
\subfloat[]{\includegraphics[height=3cm]{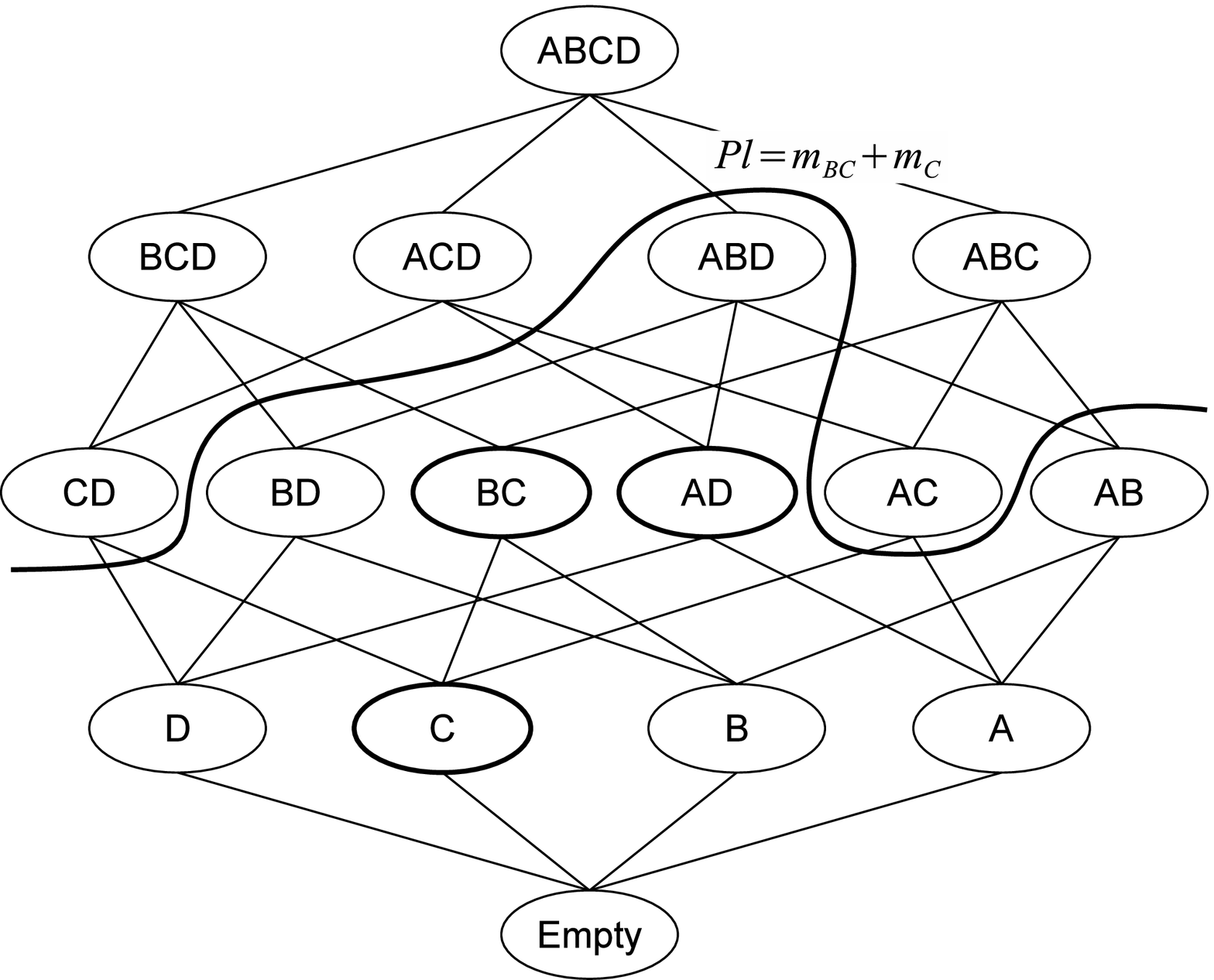}}
\hspace{4mm}
\subfloat[]{\includegraphics[height=3cm]{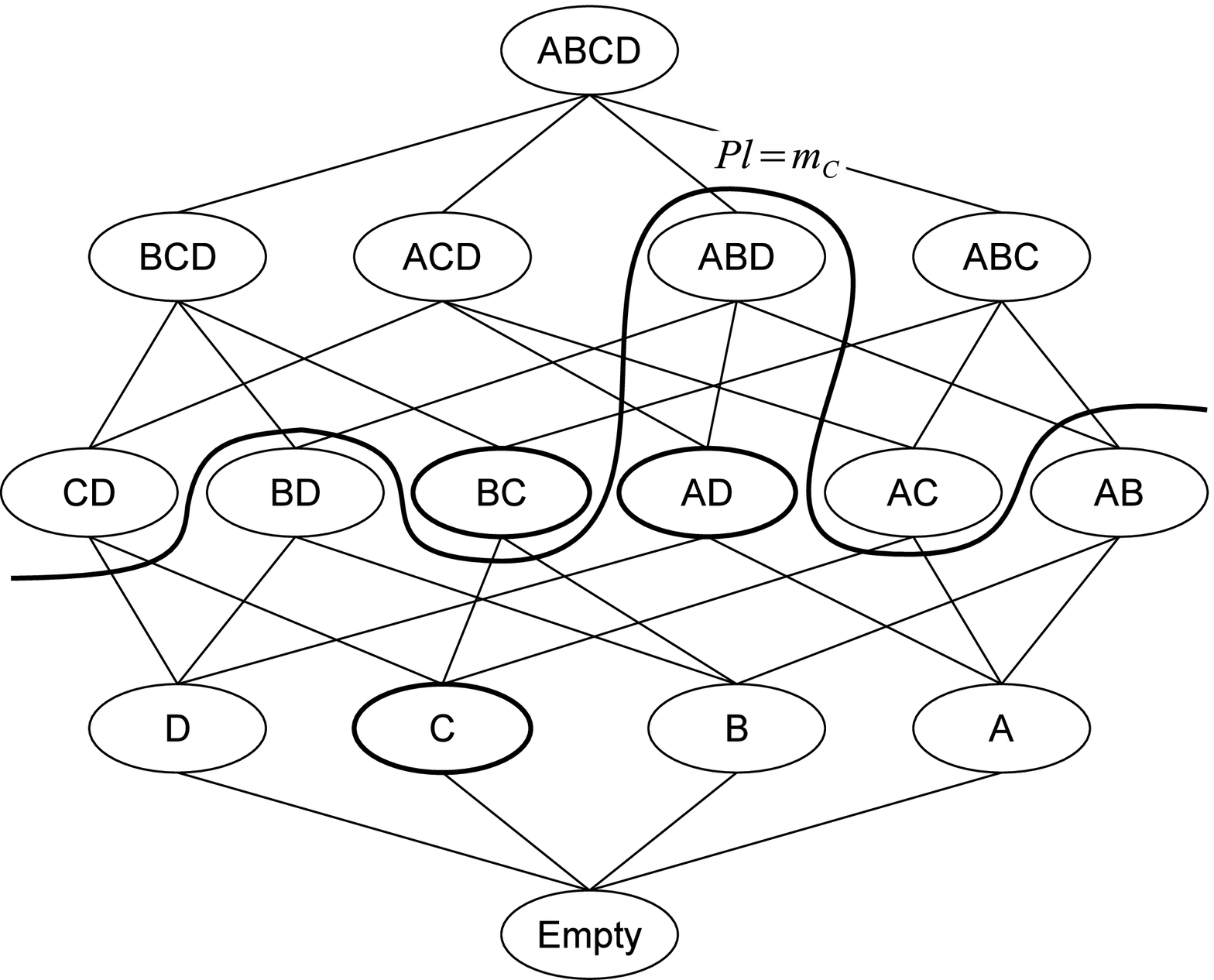}}\\

\subfloat[]{\includegraphics[height=3cm]{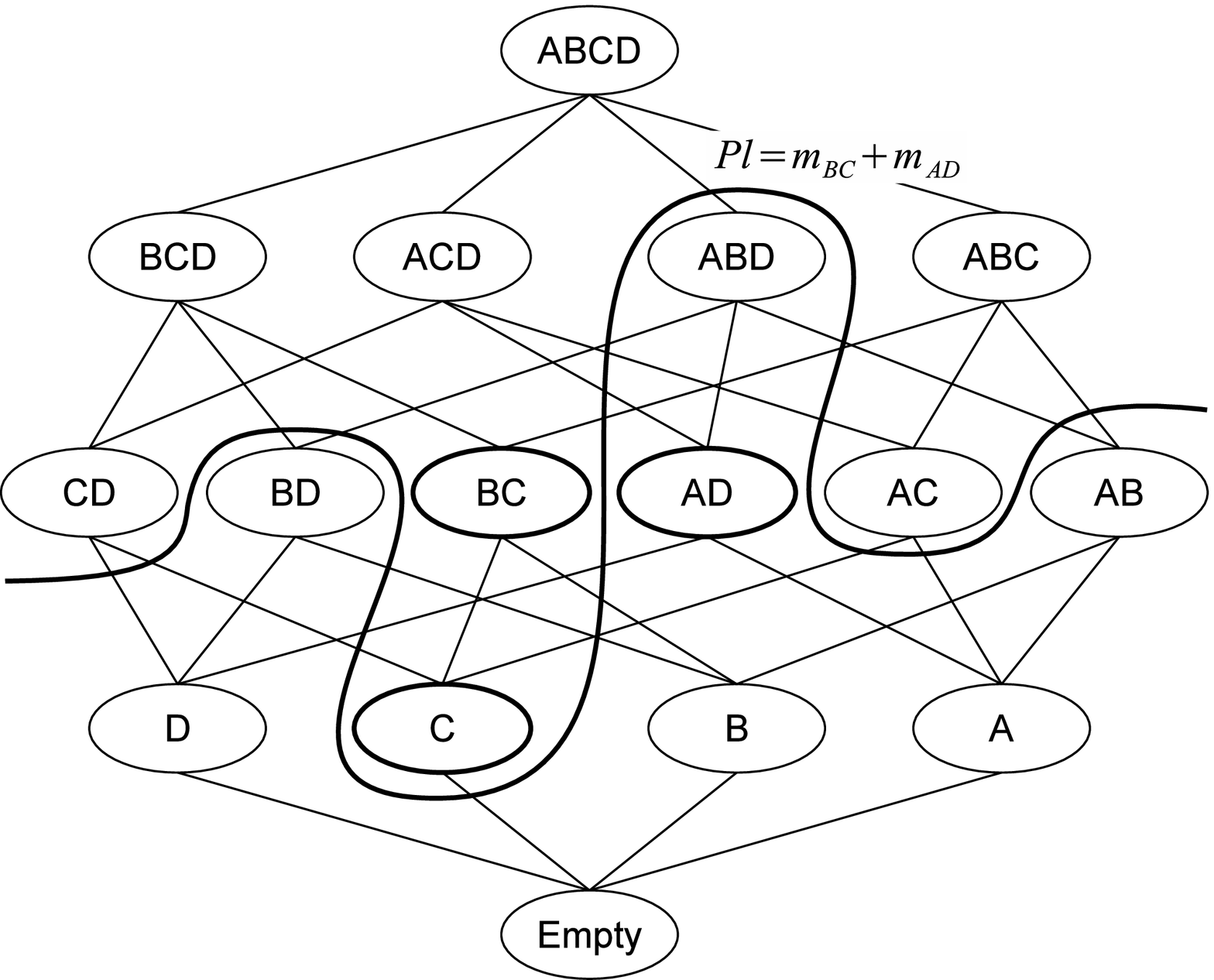}}
\hspace{4mm}
\subfloat[]{\includegraphics[height=3cm]{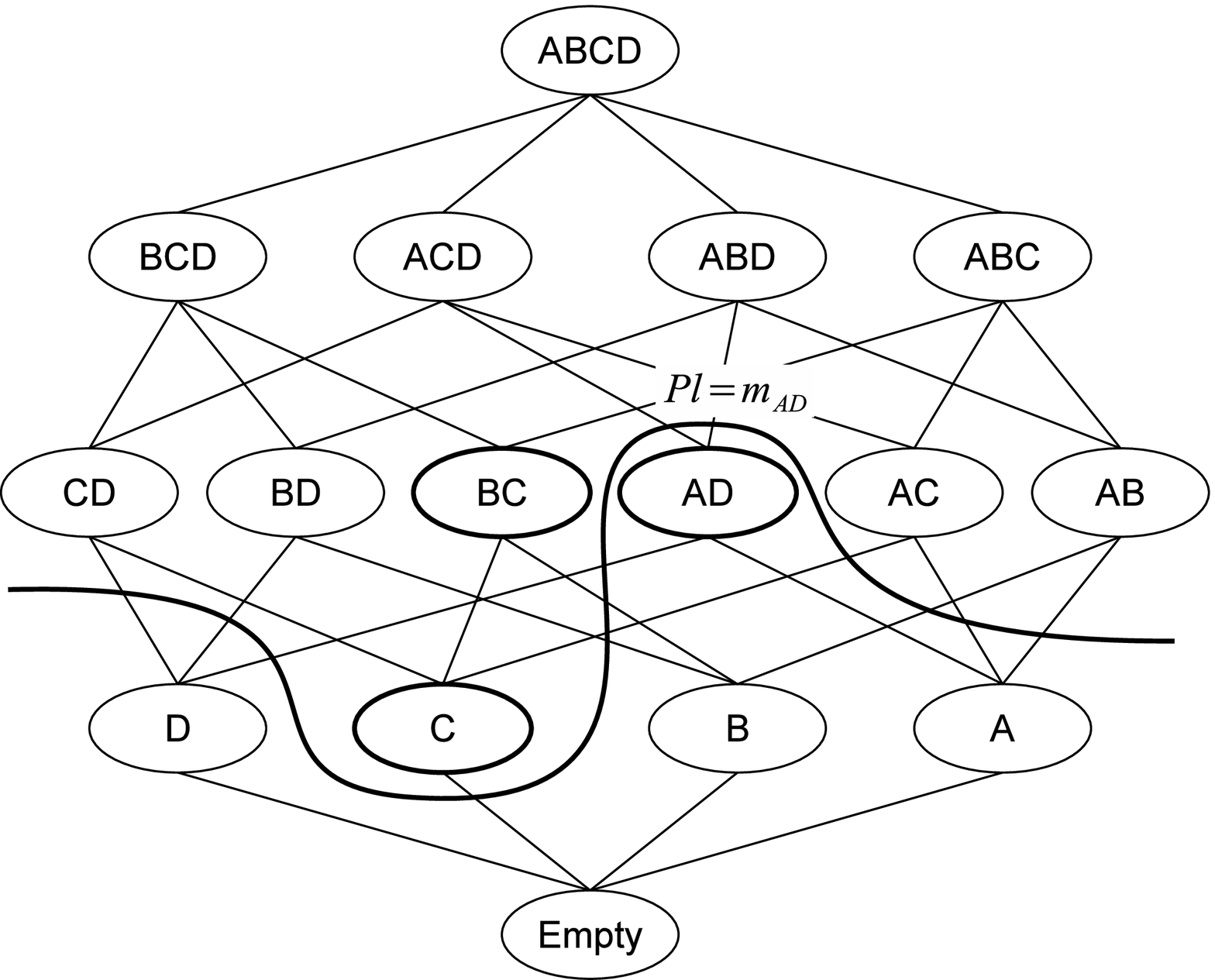}}\\
%\hspace{2mm}
\subfloat[]{\includegraphics[height=3cm]{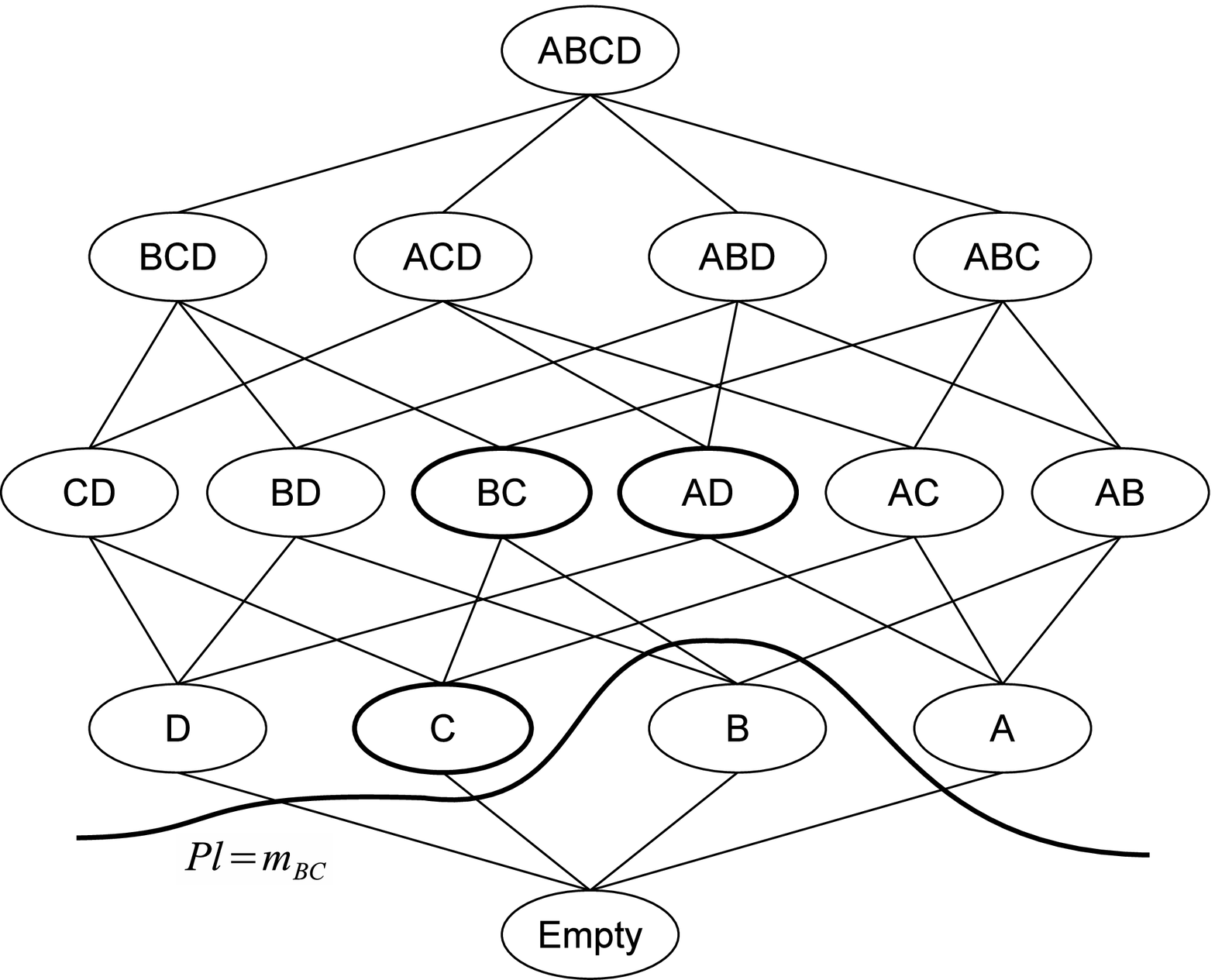}}
\hspace{4mm}
\subfloat[]{\includegraphics[height=3cm]{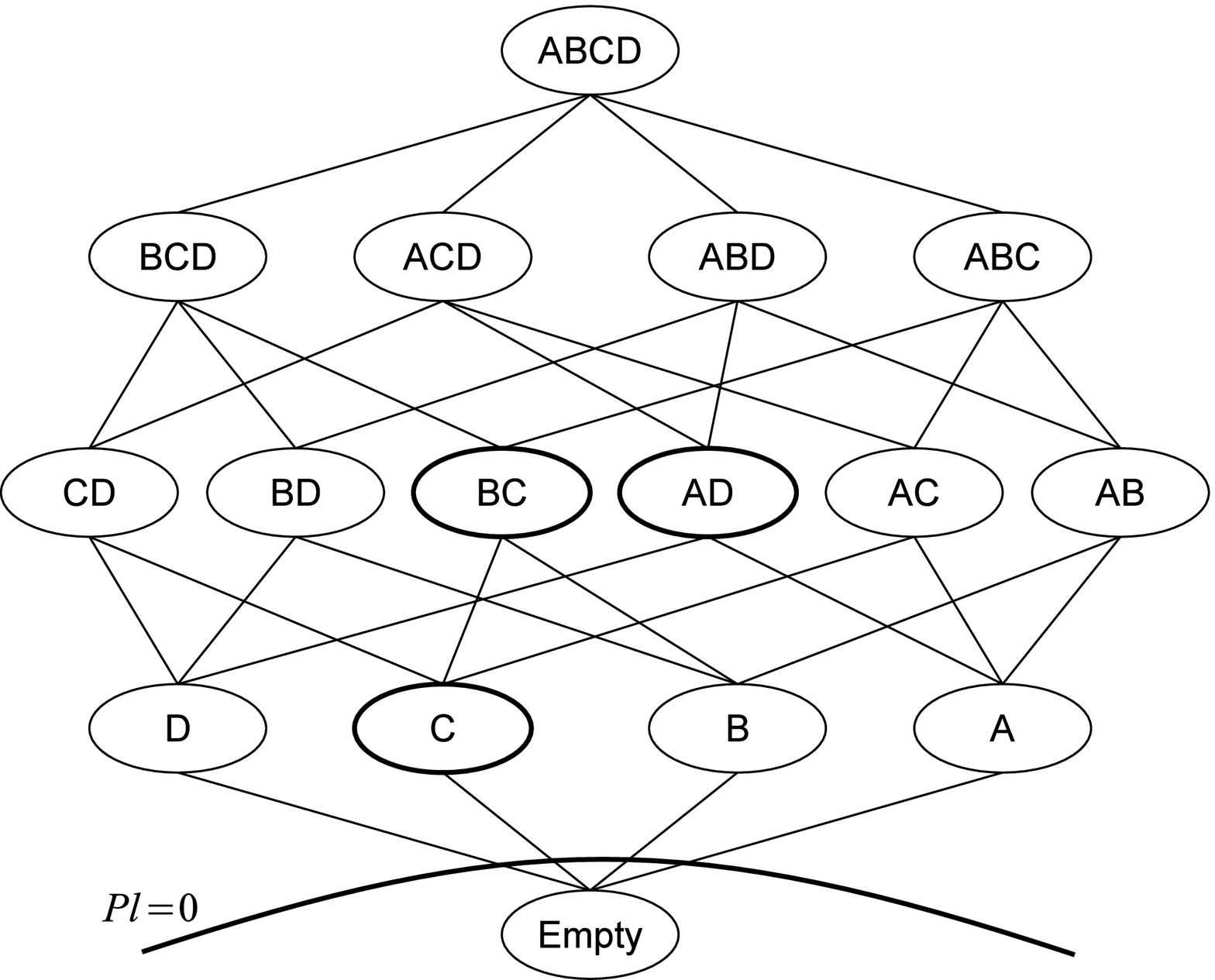}}
\caption{Plausibility levels}
\label{fig:Plausibility-levels}
\end{figure}

% \begin{figure}[t!]
% \centering
% \subfloat[]{\includegraphics[height=3cm]{lattice-bel1.eps}}
% \hspace{4mm}
% \subfloat[]{\includegraphics[height=3cm]{lattice-bel2.eps}}\\
% %\hspace{2mm}
% \subfloat[]{\includegraphics[height=3cm]{lattice-bel3.eps}}
% \hspace{4mm}
% \subfloat[]{\includegraphics[height=3cm]{lattice-bel4.eps}}\\

% \subfloat[]{\includegraphics[height=3cm]{lattice-bel5.eps}}
% \hspace{4mm}
% \subfloat[]{\includegraphics[height=3cm]{lattice-bel6.eps}}\\
% %\hspace{2mm}
% \subfloat[]{\includegraphics[height=3cm]{lattice-bel7.eps}}
% \hspace{4mm}
% \subfloat[]{\includegraphics[height=3cm]{lattice-bel8.eps}}
% \caption{Belief levels}
% \label{fig:Belief-levels}
% \end{figure}

If we sort the Belief (or Plausibility) values in ascending order, we get a sequence of levels, each grouping the nodes into those that are below the level and over the level. For instance, if we assume  
\begin{multline*}
0 \leq m_{BC} \leq m_{AD} \leq m_{BC}+m_{AD} \leq m_{C}\\
 \leq m_{BC} + m_{C} \leq m_{AD} + m_{C} \leq m_{BC} + m_{AD} + m_{C} = 1
\end{multline*}
we get the situation depicted by Fig.~\ref{fig:Plausibility-levels} with respect to Plausibility.
The following definitions enable the concept of \emph{classes of equivalence} among the subsets with respect to Belief or Plausibility and to identify those elements that are most representative of the class.

\begin{definition}[Core] Given a subset $A \subseteq \Omega$, the set of focal elements included in $A$, \emph{core} of $A$, is defined as
\begin{equation}
Cr(A) \eqdef \{ B \in F(\Omega) | B \subseteq A \}
\end{equation}    
\end{definition}

\begin{definition}[Support] Given a subset $A \subseteq \Omega$ and the set of focal elements (even partially in $A$), \emph{support} of $A$, is defined as 
\begin{equation}
Su(A) \eqdef \{ B \in F(\Omega) | B \cap A \neq \emptyset \}
\end{equation}    
\end{definition}

For instance, according to the example in Fig.~\ref{fig:lattice} $F(\Omega)=\{C, BC, AD\}$, we have $Cr(BCD)=\{C,BC\}=Cr(BC)$ and $Su(ABD)=Su(BD)=Su(AB)=\{BC,AD\}$. It is straightforward that $Cr(A) \subseteq Su(A)$, for all $A \subseteq \Omega$. The core and support represent the basis for computing respectively the Belief and the Plausibility of $A$. The core and the support are able to group the subsets of $\Omega$ into classes of equivalence as the following definition states.
 
\begin{definition}[$Cr-$ and $Su-$ Equivalence]\label{defcr} Two sets $A$ and $B$ are said to be $Cr$-equivalent if and only if $Cr(A) = Cr(B) = Cr$. A $Cr$-equivalence class is defined as the collection 
\begin{equation}
E_{Cr} \eqdef \{ A \subseteq \Omega \; | \; Cr(A) = Cr \}
\end{equation}
In addition, $A$ and $B$ are $Su$-equivalent if and only if $Su(A) = Su(B)=Su$. The $Su$-equivalence class obtained from this relation.
is defined as 
\begin{equation}
E_{Su} \eqdef \{ A \subseteq \Omega \; | \; Su(A) = Su \}
\end{equation}
\end{definition}

\begin{figure}[h!]
\centering
\subfloat[]{\includegraphics[height=3cm]{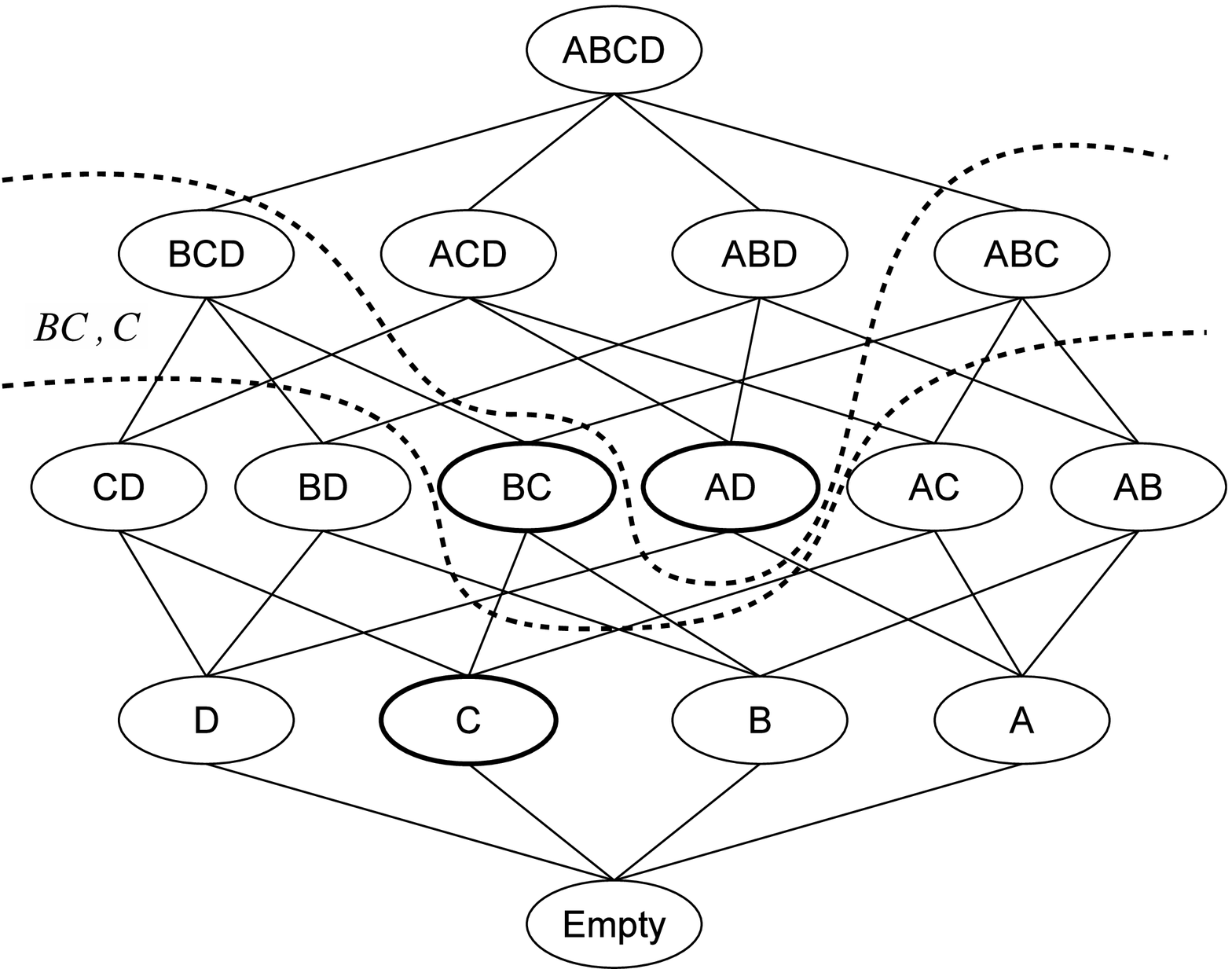}} \hspace{3mm}
\subfloat[]{\includegraphics[height=3cm]{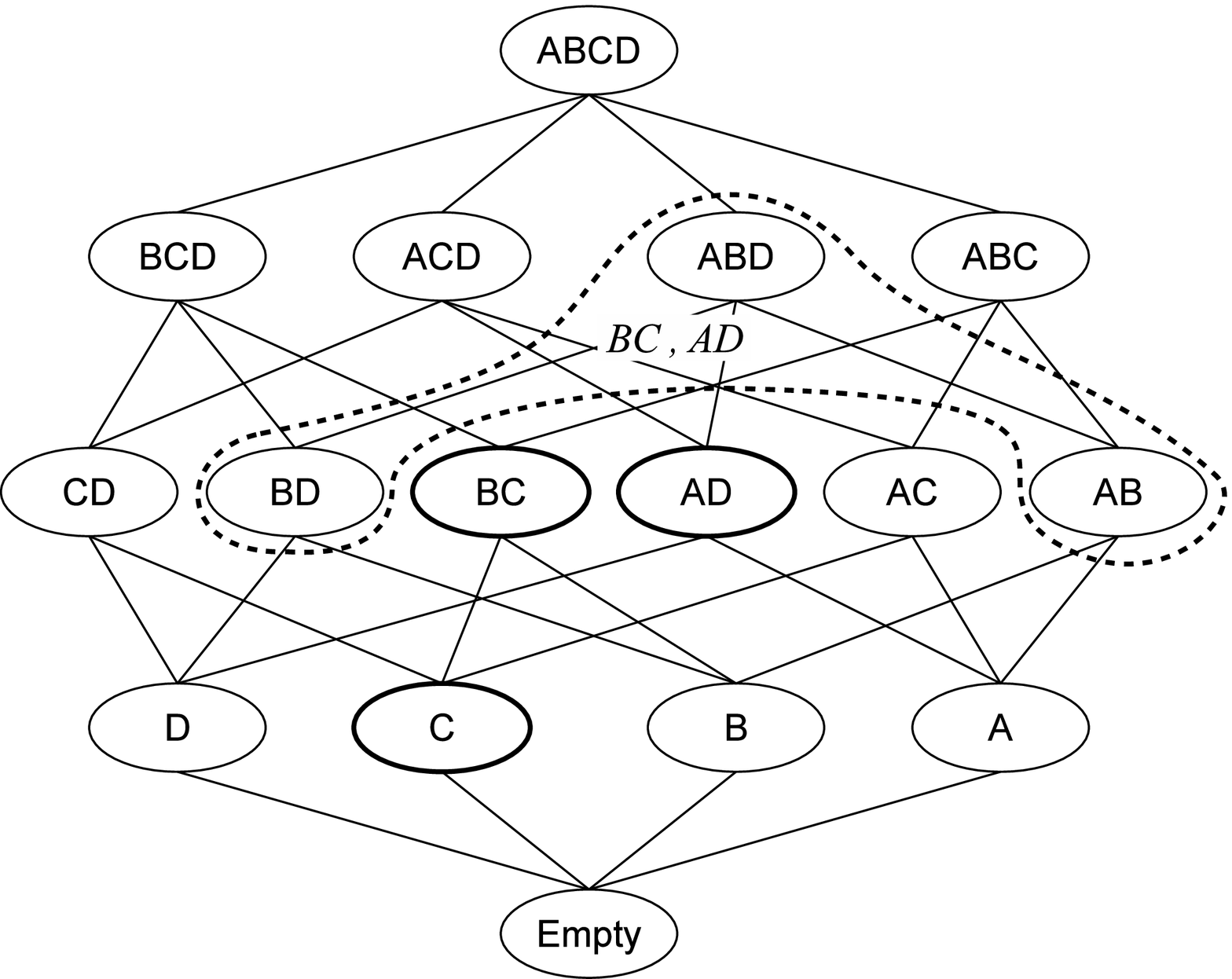}}
\caption{Examples of $Cr$-equivalence (a) and $Su$-equivalence (b) classes}
\label{fig:eq-classes}
\end{figure}

Fig.~\ref{fig:eq-classes}(a) provides an example of $Cr$-equivalence class assuming as core $Cr=\{BC,C\}$. Fig.~\ref{fig:eq-classes}(b) shows the $Su$-equivalence class for the support $Su=\{BC,AD\}$.

As an immediate consequence, if $A$ and $B$ are $Cr$-equivalent, then $Bel(A) = Bel(B)$, while if they are $Su$-equivalent, $Pl(A) = Pl(B)$.

$Cr-$ and $Su-$ equivalence classes perform a partitioning of $2^\Omega$. Thus, each subset $X \subset \Omega$ can belong only to one equivalence class. Grouping subsets in $Cr-$ and $Su-$equivalence classes allows (i) to explore the lattice by moving across classes, instead of exploring the whole item subset space, and (ii) to choose a representative of each class, so that the list of recommended items is shorter. For instance, we might be interested in using the smallest subset within a $Cr$-equivalence class.

As representative of a $Cr-$equivalence class we can assume the smallest subset. We call this set $Cr-$\emph{minimal}. For instance, for the class $\{BC, ABC, BCD\}$, the core is $\{C,BC\}$ and the $Cr-$minimal is $BC$. It is possible to prove that each $Cr-$equivalence class as one single $Cr-$minimal. Conversely, for $Su-$equivalence classes we assume as representative the largest subset, that we call $Su-$\emph{maximal}. Similarly to $Cr-$equivalence classes, it is possible to prove that any $Su-$equivalence class has one single $Su-$maximal. For example, the class $\{C,AD\}$, whose support is $\{ACD,AC,AD\}$, as $ACD$ as maximal.

\section{Model}

In the context of our interest we assume $I=\{I_1,\ldots,I_m\}$ as the set of items belonging to the content catalog, while $U=\{U_1, U_2, \dots,U_n\}$ as the set of users. 

Both sets are projected on two feature spaces, respectively made of $p$ and $q$ dimensions. The first is referred to the set of characteristics describing the items in $I$, $C=\{C_{1}, \dots , C_{p}\}$, while the second to the user profiling $P=\{P_{1},\ldots,P_{q}\}$. Both spaces are discrete, so that each $C_{i}$ and $P_{j}$ can assume a finite number of values.

The relationship between items and users is expressed by a choice matrix, as that shown in Tab.~\ref{tab:choicematrix}. The choice matrix is places side by side to the item characteristics matrix (left side) and to the profile matrix (top).

\begin{table}[!h]
	\centering
	\caption{Structure of dataset assumed by the model}
    \label{tab:choicematrix}
    \setlength{\tabcolsep}{4.5pt}
    \renewcommand{\arraystretch}{1.5}
    \def\mystrut{\rule[-2ex]{0ex}{2ex}}
    \begin{tabular}{|llll | c | ccccc |}
    \cline{5-10}
    \multicolumn{4}{c|}{}
    & $P_q$ & $p_{1,q}$ & $p_{2,q}$ & $p_{3,q}$ & $\dots$ & $p_{n,q}$ \\
    \multicolumn{4}{c|}{}
    & $P_q$ & $p_{1,q}$ & $p_{2,q}$ & $p_{3,q}$ & $\dots$ & $p_{n,q}$ \\
    \multicolumn{4}{c|}{}
    & $P_q$ & $p_{1,q}$ & $p_{2,q}$ & $p_{3,q}$ & $\dots$ & $p_{n,q}$ \\
    \multicolumn{4}{c|}{}
    & $\vdots$ & $\vdots$ & $\vdots$ & $\vdots$ & $\ddots$ & $\vdots$ \\
    \multicolumn{4}{!{\mystrut}c|}{}
    & $P_q$ & $p_{1,q}$ & $p_{2,q}$ & $p_{3,q}$ & $\dots$ & $p_{n,q}$ \\     
    \hline
    $C_1$ & $C_2$ & \dots & $C_ p$ & \backslashbox{$I$}{$U$} & $1$ & $2$ & $3$ & \dots & $n$ \\ 
    \hline
    $c_{1,1}$    & $c_{1,2}$ &  $\dots$& $c_{1,p}$ 
    & 1   & $\checkmark$ & $\checkmark$ & ~ & $\dots$ & ~ \\
    $c_{2,1}$    & $c_{2,2}$ &  $\dots$& $c_{2,p}$ 
    & 2   & $\checkmark$ & ~ & $\checkmark$ & $\dots$ & ~ \\
    $c_{3,1}$    & $c_{3,2}$ &  $\dots$& $c_{3,p}$ 
    & 3   & ~ & $\checkmark$ & $\checkmark$ & $\dots$ & $\checkmark$ \\
    $\vdots$ & $\vdots$ & $\ddots$ & $\vdots$ &
    $\vdots$ & $\vdots$ & $\vdots$ & $\vdots$ & $\ddots$ & $\vdots$ \\
    $c_{m,1}$    & $c_{m,2}$ &  $\dots$& $c_{m,p}$
    & m  & $\checkmark$ & $\checkmark$ & ~ & $\dots$ & ~     \\
    \hline
    \end{tabular}
\end{table}

In general, data points $c_{i,h}$ and $p_{j,k}$ are multi-valued, meaning that they are represented by sets of values. For instance if $C_h$ is representing the movie cast, $c_{i,h}$ is represented by the list of actors that are featuring in the movie $I_i$. Similarly, if $P_k$ is "interests", $p_{j,k}$ will list what the user $U_j$ is interested in. In other cases they are single-valued, such as in the case of characteristics such as "director" and "year" or in the case of profiling features such as "age" or "location". An example of this matrix is given in Tab.\ref{tab:dataset_ex}.

\begin{table*}[h!]
	\centering
    
	\caption{The dataset used as example.}
    \label{tab:dataset_ex}
    
    \setlength{\tabcolsep}{4.5pt}
    \renewcommand{\arraystretch}{1.5}

\scalebox{0.95}[0.95]{
    \begin{tabular}{|llll|c|P{1cm}P{1cm}P{1cm}P{1cm}|}
    
    \cline{5-9}
    \multicolumn{4}{c|}{}
    & Age & 30s & 30s & 20s & 40s \\
    %\cdashline{5-9}
    \multicolumn{4}{c|}{}
    & Gender & M & F & M & M \\
    %\cdashline{5-9}
    \multicolumn{4}{c|}{}
    & Location & IT & IT & SP & IT \\
    \cdashline{5-9}
    \multicolumn{4}{c|}{}
    & Interests & Movies Books & Sport & Books & Music Sport \\
    
    \hline
       Director & Year & Stars & Genre & \backslashbox{$I$}{$U$} & 1 & 2 & 3 & 4 \\
	\hline
       Boyle  & 1996  
       & Ewan McGregor, Ewen Bremner
       & Drama
       & 0 & $\checkmark$ & $\checkmark$ & $\checkmark$ &   \\
       
       Levinson  & 1996  
       & Robert De Niro, Kevin Bacon, Brad Pitt 
       & Crime, Drama, Thriller 
       & 1 & $\checkmark$ & & & $\checkmark$ \\

       Scorsese  & 2015  
       & Robert De Niro, Leonardo DiCaprio, Brad Pitt 
       & Short, Comedy 
       & 2 & & $\checkmark$ & & \\
 
 	   Scorsese  & 1990  
       & Robert De Niro, Ray Liotta, Joe Pesci 
       & Biography, Crime, Drama
       & 3 & & & $\checkmark$ & \\
 
  	   Boyle  & 2000  
       & Leonardo DiCaprio
       & Adventure, Drama, Romance 
       & 4 & & & $\checkmark$ & \\
       
       Howard  & 1995  
       & Tom Hanks, Kevin Bacon 
       & Adventure, Drama, History 
       & 5 & & $\checkmark$ & & $\checkmark$ \\
       
       Zemeckis  & 1994  
       & Tom Hanks
       & Comedy, Drama 
       & 6 & $\checkmark$ & & & \\
       
       Zemeckis  & 1985  
       & Michael J. Fox, Christopher Lloyd
       & Adventure, Sci-Fi 
       & 7 & & & & $\checkmark$ \\  
       
       Edwards  & 2016  
       & Felicity Jones, Diego Luna
       & Adventure, Sci-Fi 
       & 8 & & $\checkmark$ & $\checkmark$ & \\
       
       Scott  & 2015 
       & Matt Damon
       & Adventure, Drama, Sci-Fi
       & 9 & & $\checkmark$ & & \\       
       
    \hline
    \end{tabular}
}
\end{table*}

Let us denote with $\Phi(C_h)$ the overall set of values assumed over the item characteristic $C_h$, and with $\Phi(P_k)$ the overall set of values for the user profiling feature $P_k$. They are respectively given in Tab.\ref{tab:char_sets} and Tab.\ref{tab:prof_sets}.

\begin{table}[!h]
	\centering
	\caption{Overall sets of item characteristics}
    \label{tab:char_sets}
    \begin{tabular}{|c:c:c:c|}
    \hline
    Director & Year & Actors & Genre \\
    \hline
    Boyle    & 1996 & Ewan McGregor     & Crime     \\
    Levinson & 2015 & Ewen Bremner      & Drama     \\
    Scorsese & 1990 & Ray Liotta        & Thriller  \\
    Howard   & 2000 & Robert De Niro    & Short     \\
    Zemeckis & 1995 & Kevin Bacon       & Comedy    \\
    Edwards  & 1994 & Brad Pitt         & Biography \\
    Scott    & 1985 & Leonardo DiCaprio & Adventure \\
             & 2016 & Joe Pesci         & Romance   \\
             &      & Ray Liotta        & History   \\
             &      & Tom Hanks         & Sci-Fi    \\
             &      & Michael J. Fox    &           \\
             &      & Christopher Lloyd &           \\
             &      & Felicity Jones    &           \\
             &      & Diego Luna        &           \\
             &      & Matt Damon        &           \\
    \hline
	\end{tabular}
\end{table}

\begin{table}[!h]
	\centering
	\caption{Overall sets of user profiling features}
    \label{tab:prof_sets}
    \begin{tabular}{|c:c:c:c|c:c:c:c|}
    \hline
    Age & Gender & Location & Interests \\
    \hline
    20s & M & IT & Books \\
    30s & F & SP & Movies \\
    40s &   & SP & Sport \\
        &   &    & Music \\ 
    \hline       
	\end{tabular}
\end{table}

Since here we are interested to use both information regarding the item characteristics and the user profiles, we compute for any
\begin{equation}
m(K) = \frac{|L(K)|}{|L|}
\end{equation}
where
\begin{itemize}
\item $K \subseteq \Phi$, with $\Phi$ being the overall set of a given characteristic $C_h$ or a profiling feature $P_k$.  
\item $L \subseteq I \times U$ is the set of preferences ("likes")
\item $L(K) \subseteq L$ is the subset of preferences referred to $K$ 
\end{itemize}
It is easy to prove that $m(\emptyset) = 0$ and $m(\Phi) = 1$. Assuming that in our example $|L| = 15$, some example of masses assigned to characteristics are given below.
\begin{itemize}
\item Stars: $m(De\ Niro, Bacon, Pitt) = 	\frac{2}{15}$
\item Director: $m(Boyle) = \frac{4}{15}$
\item Year: $m(1996) = \frac{5}{15}$
\item Genre: $m(Drama) = \frac{3}{15}$
\end{itemize}
If we refer to profiling features, some examples are the following:
\begin{itemize}
\item Age: $m(30s) = \frac{8}{15}$
\item Gender: $m(F) = \frac{5}{15}$
\item Location: $m(IT) = \frac{11}{15}$
\item Interests: $m(Movies, Books) = \frac{3}{15}$
\end{itemize}

We notice that focal elements of each dimension are given by its unique values, i.e. by rows after removing duplicates. For instance, for the "Director" and "Year" dimensions, focal elements are given by the set of director names, i.e., Boyle, Levinson, Scorsese, Howard, Zemeckis, Edwards and Scott, and by years, i.e., 1985, 1990, 1994, 1995, 1996, 2000, 2015, 2016. Similarly for "Age", i.e., 20s, 30s, 40s, "Location", i.e., IT, SP, and "Gender", i.e., M, F. They are all single-value dimensions. For them, focal elements are singletons. In this case the model becomes additive. For instance,
\begin{itemize}
\item $Bel(Scorsese, Boyle) = m(Scorsese) + m(Boyle)$
\item $Pl(Scorsese, Boyle) = m(Scorsese) + m(Boyle)$
\end{itemize}

Instead, the dimensions "Genre" and "Stars" are multi-value, so their focal elements are not singletons. For instance, Drama, Comedy-Drama and  Adventure-Drama-History are three focal elements of "Genre". Similarly, Movies-Books, Books, Sport, Music-Sport are focal elements of "Interests" among the profiling features. For multi-value dimensions the model is not additive. As an example, let us consider the belief of Adventure-Comedy-Sci-Fi-Drama. We have,
\begin{equation*}
\begin{split}
Bel( Adventure, Comedy, Sci-Fi, Drama ) = \\
m(Adventure, Drama, Sci-Fi) + \\
m(Adventure, Sci-Fi) + \\
m(Comedy, Drama) + \\
m(Drama) = \\
\frac{1}{15} + \frac{3}{15} + \frac{1}{15} + \frac{3}{15} = \frac{8}{15}
\end{split}
\end{equation*}
Conversely, we have
\begin{equation*}
Pl( Adventure, Comedy, Sci-Fi, Drama ) = 1\\
\end{equation*}
because all focal elements of "Genre" are involved in its computation.

So far, we considered each dimension in isolation. They provide a range for probability $Pr(K) = [Bel(K), Pl(K)]$, with $K \subseteq \Phi$,  that is a measure of likelihood that a content in $I$ characterized by $K$ will be enjoyed by the set of users in $U$, if $\Phi$ is referred to some item characteristic $C_h$.  Or, if we look at $K$ as referred to some profiling feature $P_k$, it is the likelihood that a user in $U$ will enjoy the catalog of contents offered by means of $I$.
 
If we would like to look at multiple dimensions we are not allowed to use the Dempter's combination rule as described in the section above. The main issue is that dimensions belong to different domains, so that the information fusion given by Eq.\eqref{eq:dempster_rule} cannot be performed over comparable sets. This problem can be solved when we look at focal elements as representative of preferences over the matrix $L$. Let $K_1$ and $K_2$ two features defined over different dimensions. We can combine the two by means of conjunction or disjunctions, depending on the semantics we associate to the operation.

Thus, in order to perform a combination of $K_1$ and $K_2$ we need to look at $L(K_1)$ and $L(K_2)$. In the case of conjunction $K = K_1 \odot K_2$, we have $L(K) = L(K_1) \cap L(K_2)$, so that
\begin{equation}
m(K) = \frac{|L(K_1) \cap L(K_2)|}{|L|}
\end{equation}

For instance, if $K_1$ is Zemeckis and $K_2$ is Adventure-Sci-Fi, we have that $L(K1)$ is made of preferences at rows 6 and 7, while $L(K_2)$ at rows 7 and 8, so that $L(K_1) \cap L(K_2)$ is made only of row 7, and $m(K) = \frac{1}{15}$. Once we have focal elements for the conjunction of the "Director" and "Genre", we can compute the belief and plausibility over the conjunction of the two. For instance,
\begin{equation*}
Bel(Zemeckis \odot Drama) = m(Zemeckis \odot Drama) = \frac{1}{15} 
\end{equation*}
and
\begin{equation*}
Pl(Zemeckis \odot Drama) = m(Zemeckis \odot Drama) = \frac{1}{15} 
\end{equation*}

The meaning of $Pr(Zemeckis \odot Drama)$ is the likelihood that a drama directed by Zemeckis will be enjoyed given $L$ and users in $U$, that is exactly 1 over 15. 

The other way of combining two dimensions is by means of disjunction. In this case $K = K_1 \oplus K_2$ and
\begin{equation}
m(K)= \frac{|L(K_1) \cup L(K_2)|}{|L|} 
\end{equation}
For instance, with regard to profiling features, if $K_1$ is 20s and $K_2$ is Sport, we have
\begin{equation*}
Bel(Sport \odot 20s) = \frac{7}{15} 
\end{equation*}
as the conjunction of the two collect rows 0,2,3,4,5,8,9. In this case, both belief and plausibility are larger.

\section{Applications in Media Industry}

The model presented so far can be employed for different tasks. We briefly outline some of them below.

\textbf{Recommendations.} The model can be used to suggest a content to a user according to each dimension. For instance, chosen the dimension of "Director", the system might suggest directors that are most likely be of interest for the user. It is also possible to combine different dimensions. For example, "Genre" and "Year". In any case, the inference of preferences is performed by looking at users indistinguishably, meaning that profile information is not taken into account.

\textbf{Audience targeting.} In this case, given a single content we are interested to find user profiles that might be interested to it. For example, given a new movie, the model might estimate how likely could be of interest for each range of age. Also in this case it is possible to combine multiple dimensions that are user profiling features. For instance, considering multiple age ranges, taking into account the different genders.

\textbf{Content bundling.} This application is aimed to propose a bundle of contents to a group of users, possibly with different profiles. This result can be suggested by the model through a combination of dimensions among characteristics and profiling features. The process can be led by two different perspectives. The first moves from the bundle of contents and it is aimed at identifying a group of users that might be interested in. For instance, given all drama movies in 90s, which users could be interested to such an offer. But it is also possible to move the other way round: selected a group of users, what is the bundle of contents that might be of their interest. With respect to our example, given users in the 20s that are interested to books, what is the bundle of contents that could be likely of their interest. 

\textbf{Segmentation.} This is a generalization of the problem above. In this case both users and contents are objective of the analysis. We are interested to find clusters of users, contents and user/contents that maximize the likelihood of preferences within the group and minimize the likelihood of preferences between groups. For instance, by looking at our example, we could be interested to see if there are users with different profiles that are likely to enjoy the same contents, or if there are contents are have similar likelihood to be enjoyed by the audience, or if there groups of users that are likely to enjoy the same group of contents, besides the others.

In all tasks above, it plays a key role the possibility of comparing and ranking alternatives. However, the D-S theory provides only an imprecise probability that ranges between the lower bound given by the degree of belief and the upper bound given by the degree of plausibility. This issue can be addressed by different approaches. 

The first approach is to use a degree that is representative of a range, such as the middle point between belief and plausibility. Another possibility is to use only belief degrees (conservative approach) or plausibility (challenging approach). Another approach could be to randomly choose $n$ pairs from both ranges and to use the majority or pairwise comparisons in order to decide the order of two alternatives. It is also possible to choose randomly an alternative when the two cannot be sorted. Finally, it is possible to look at other solutions investigated in the field of partial order theory.

\section{Conclusions}

In this paper, we further investigated a preference model based on the Dempster-Shafer theory and its application to media industry. This work is an evolution of what has been done so far by introducing some elements of novelty. Among them the possibility of including the user profile as part of the inference, instead of being considered neutrally with respect to different applications and problems that have been discussed in the section before. There are still some issue to address. The most important is referred to scalability of the model. Indeed, the nature of the D-S theory is inherently combinatorial, so that the search space is exploding by including more elements within the dimension overall sets $\Phi$. The possibility of defining equivalence classes in terms of belief and plausibility is a way to reduce complexity, but still work has to be done to make this solution feasible in practice. In addition, the model presented here requires to be validated. This can be done by looking at correspondences between the probability ranges and the frequency of positive voted that are after recorded. In the future we aim to develop further the model in order to include more complex queries and to solve issues regarding the application of the model in practice with respect to large catalogs and audience.

%\footnotesize
%\vspace{5mm}
%\noindent
%\textbf{DISCLAIMER. This article was prepared or %accomplished by authors in their personal capacity. The opinions expressed in this paper are the authors' own and do not reflect the view of Sky Italia.}


%\section*{Acknowledgment}



%\bibliographystyle{IEEEtran}

\bibliographystyle{myIEEETran}

%\bibliography{references.bib}
\bibliography{IEEEabrv,references.bib}

\begin{thebibliography}{10}
\providecommand{\url}[1]{#1}
\csname url@samestyle\endcsname
\providecommand{\newblock}{\relax}
\providecommand{\bibinfo}[2]{#2}
\providecommand{\BIBentrySTDinterwordspacing}{\spaceskip=0pt\relax}
\providecommand{\BIBentryALTinterwordstretchfactor}{4}
\providecommand{\BIBentryALTinterwordspacing}{\spaceskip=\fontdimen2\font plus
\BIBentryALTinterwordstretchfactor\fontdimen3\font minus
  \fontdimen4\font\relax}
\providecommand{\BIBforeignlanguage}[2]{{%
\expandafter\ifx\csname l@#1\endcsname\relax
\typeout{** WARNING: IEEEtran.bst: No hyphenation pattern has been}%
\typeout{** loaded for the language `#1'. Using the pattern for}%
\typeout{** the default language instead.}%
\else
\language=\csname l@#1\endcsname
\fi
#2}}
\providecommand{\BIBdecl}{\relax}
\BIBdecl

\bibitem{Gomez2015}
\BIBentryALTinterwordspacing
C.~A. Gomez-Uribe and N.~Hunt, ``The netflix recommender system: Algorithms,
  business value, and innovation,'' \emph{ACM Trans. Manage. Inf. Syst.},
  vol.~6, no.~4, pp. 13:1--13:19, Dec. 2015. doi: 10.1145/2843948. [Online].
  Available: \url{http://doi.acm.org/10.1145/2843948}
\BIBentrySTDinterwordspacing

\bibitem{Salter2006}
\BIBentryALTinterwordspacing
J.~Salter and N.~Antonopoulos, ``Cinemascreen recommender agent: Combining
  collaborative and content-based filtering,'' \emph{IEEE Intelligent Systems},
  vol.~21, no.~1, pp. 35--41, Jan. 2006. doi: 10.1109/MIS.2006.4. [Online].
  Available: \url{http://dx.doi.org/10.1109/MIS.2006.4}
\BIBentrySTDinterwordspacing

\bibitem{Candillier_comparingstate-of-the-art}
L.~Candillier, F.~Meyer, and M.~Boull\'e, ``Comparing state-of-the-art
  collaborative filtering systems,'' pp. 548--562, 2007.

\bibitem{Pazzani1999}
\BIBentryALTinterwordspacing
M.~J. Pazzani, ``A framework for collaborative, content-based and demographic
  filtering,'' \emph{Artif. Intell. Rev.}, vol.~13, no. 5-6, pp. 393--408, Dec.
  1999. doi: 10.1023/A:1006544522159. [Online]. Available:
  \url{http://dx.doi.org/10.1023/A:1006544522159}
\BIBentrySTDinterwordspacing

\bibitem{Burke00knowledge-basedrecommender}
R.~Burke, ``Knowledge-based recommender systems,'' in \emph{Encyclopedia of
  Library and Information Science, vol. 69}, A.~Kent, Ed.\hskip 1em plus 0.5em
  minus 0.4em\relax Taylor and Francis, 2000, pp. 180--201.

\bibitem{Bobadilla2013109}
\BIBentryALTinterwordspacing
J.~Bobadilla, F.~Ortega, A.~Hernando, and A.~Guti{\'e}rrez, ``Recommender
  systems survey,'' \emph{Knowledge-Based Systems}, vol.~46, no.~0, pp. 109 --
  132, 2013. doi: 10.1016/j.knosys.2013.03.012. [Online]. Available:
  \url{http://www.sciencedirect.com/science/article/pii/S0950705113001044}
\BIBentrySTDinterwordspacing

\bibitem{CastroSchez20112441}
J.~J. Castro-Schez, R.~Miguel, D.~Vallejo, and L.~M. L\'opez-L\'opez, ``A
  highly adaptive recommender system based on fuzzy logic for \{B2C\}
  e-commerce portals,'' \emph{Expert Systems with Applications}, vol.~38,
  no.~3, pp. 2441 -- 2454, 2011. doi: 10.1016/j.eswa.2010.08.033

\bibitem{Zhang10}
K.~Zhang and H.~Li, ``Fusion-based recommender system,'' in \emph{Information
  Fusion (FUSION), 2010 13th Conference on}, July 2010. doi:
  10.1109/ICIF.2010.5712091 pp. 1--7.

\bibitem{Troiano201598}
L.~Troiano, L.~J. Rodr{\'i}uez-Mu{\~n}iz, and I.~D{\'i}az, ``Discovering user
  preferences using dempster-shafer theory,'' \emph{Fuzzy Sets and Systems},
  vol. 278, pp. 98 -- 117, 2015. doi:
  http://dx.doi.org/10.1016/j.fss.2015.06.004

\bibitem{dempster67a}
A.~P. Dempster, ``{Upper and lower probabilities induced by a multivalued
  mapping},'' \emph{Annals of Mathematical Statistics}, vol.~38, pp. 325--339,
  1967.

\bibitem{shafer1976mathematical}
G.~Shafer, \emph{A Mathematical Theory of Evidence}.\hskip 1em plus 0.5em minus
  0.4em\relax Princeton: Princeton University Press, 1976.

\end{thebibliography}

\end{document}